%% file: MAIN_DRAFT.tex
\documentclass[sigconf]{acmart}

\usepackage{enumitem}
\AtBeginDocument{%
  }

\copyrightyear{2026}
\acmYear{2026}
\setcopyright{cc}
\setcctype{by}
\acmConference[WWW '26]{Proceedings of the ACM Web Conference 2026}{April 13--17, 2026}{Dubai, United Arab Emirates}
\acmBooktitle{Proceedings of the ACM Web Conference 2026 (WWW '26), April 13--17, 2026, Dubai, United Arab Emirates}
\acmPrice{}
\acmDOI{10.1145/3774904.3792901}
\acmISBN{979-8-4007-2307-0/2026/04}

\begin{document}

\title{Beyond Playtesting: A Generative Multi-Agent Simulation System for Massively Multiplayer Online Games}

\author{Ran Zhang}
\authornote{Both authors contributed equally to this research. \textsuperscript{\textdagger} Corresponding author.}
\author{Kun Ouyang}
\authornotemark[1]
\affiliation{%
  \institution{LIGHTSPEED}
  \country{Singapore}
}
\email{rzhang008@e.ntu.edu.sg}
\email{ouyangkun@u.nus.edu}

\author{Tiancheng Ma}
\author{Yida Yang}
\affiliation{%
  \institution{National University of Singapore}
  \institution{Nanyang Technological University}
  \country{Singapore}
}
\email{maktubchn@gmail.com}
\email{yangyida03@gmail.com}

\author{Dong Fang\textsuperscript{\textdagger}}
\affiliation{%
  \institution{LIGHTSPEED}
  \country{Singapore}
}
\email{df572@outlook.com}

\renewcommand{\shortauthors}{Ran Zhang, Kun Ouyang, Tiancheng Ma, Yida Yang, and Dong Fang}

\input{Sections/Abstract_oy}



\maketitle

\input{Sections/Introduction_short}

\input{Sections/System}
\input{Sections/Method}

\input{Sections/Experiment}

\bibliographystyle{ACM-Reference-Format}
\bibliography{Bibs/sample-base}

\end{document}

%% file: Sections/Abstract_oy.tex
\begin{abstract}

Optimizing numerical systems and mechanism design is crucial for enhancing player experience in Massively Multiplayer Online (MMO) games. Traditional optimization approaches rely on iterative online experiments or parameter tuning over abstracted statistical models, which can be inaccurate, time-consuming and potentially impair players' experience. Although simplified offline simulation systems are frequently employed as alternatives, their low fidelity constrains agents’ ability to faithfully replicate real players’ reasoning processes and behavioral responses to interventions. To address these limitations, we propose a generative agent-based MMO simulation system with hundreds of agents empowered by Large Language Models (LLMs). By applying Supervised Fine-Tuning (SFT) and Reinforcement Learning (RL) on large-scale real player behavioral data, we adapt LLMs from general priors to game-specific domains, enabling realistic and interpretable player decision-making. In parallel, a data-driven environment model trained on real gameplay logs reconstructs dynamic in-game systems. Experiments demonstrate strong consistency with real-world player behaviors and plausible causal responses under interventions, providing a reliable, interpretable, and cost-efficient framework for data-driven numerical design optimization.

\end{abstract}

%% file: Sections/Introduction_short.tex
\section{Introduction}
Massively Multiplayer Online (MMO) games hold a significant share in the global gaming market with continuously growing player bases and profit. In complex MMO games, numerical systems and mechanism design are the two foundational pillars of gameplay. Mechanism design establishes the possibility space and fundamental interaction rules. In contrast, numerical systems precisely tune the learning curve, cost, and rewards of these interactions. Together, they are essential for shaping the player experience, maintaining balance, and ensuring long-term game sustainability.

A good example of these dynamics is seen in the emerging \textit{extraction shooter} MMO games (e.g., \textit{Escape from Tarkov}). In these games, players undertake missions in hazardous environments to acquire valuable loot and must successfully extract to secure their collected items. A key challenge is to design a mechanism balancing resource/currency accumulation and consumption. An unhealthy economic system inevitably faces severe inflation, erodes player motivation and increases churn rate. However, game economics numerical design is far from easy. In-game Parameter tuning such as currency output, production rates, or tax policies often triggers non-linear "Domino effects". Traditional multi-parameter tuning relies heavily on empirical settings, followed by a slow, iterative cycle of \textit{tuning–observe–retuning}. This approach presents significant technical and operational drawbacks, underscoring the critical need for simulation methods: a. \textit{High Time Costs}: Evaluating adjustments requires observation periods of weeks or months, a temporal challenge limiting timely feedback; b. \textit{High Opportunity Costs}: Incorrect adjustments risk irreversible negative impacts on the game's economy and player base, potentially leading to widespread user churn and long-term damage. c. \textit{Testing Limitations}: Major mechanism changes (such as introducing trading systems or opening new gameplay features) cannot be validated through small-scale A/B testing. Numerical simulation methods based on statistical models can be adopted as alternatives. Yet, it operates as a "black-box" and yields only macroscopic predictions, critically lacking the valuable microscopic, player-level insight necessary for game design.
\begin{figure*}[tbp]
  \centering
  \includegraphics[width=\textwidth]{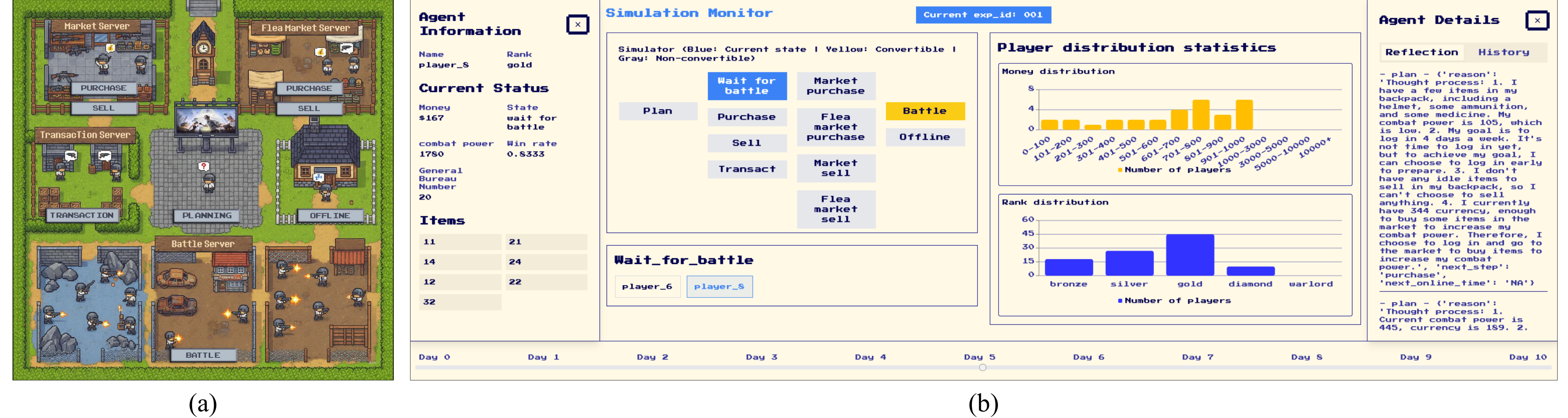}
  \vspace{-4ex}
  \caption{(a) Simulated MMO game world with multiple generative agents acting as players; (b) Simulation monitor front-end for macroscopic and microscopic monitoring of player agents.}
  \Description{Simulated MMO game}
  \vspace{-4ex}
  \label{fig:gui}
\end{figure*}

Recent advances in Large Language Models (LLMs) have spawned a new research direction utilizing generative agents to simulate virtual society. For example, AgentSociety \cite{agentsociety} and Stanford AI Town \cite{park2023generative} utilize LLM-driven agents to simulating agents v.s. agents and agents v.s. environment interactions, and observe various phenomena and mechanisms analogous to those found in human societies. ChatDev \cite{qian2023communicative} examined how agents collaborate to complete software engineering tasks; CompeteAI \cite{zhao2023competeai} studied competition dynamics between agents within communities, successfully validating the Matthew Effect. All the works demonstrate the general understanding and logical capabilities of LLM-empowered agents (generative agents), exhibiting human-like perception, reasoning, and decision-making behaviors. 

However, research addressing game simulation systems remains extremely limited. While recent works, such as Xu et al.'s exploration of player negotiation via generative agents \cite{NetEase_empowering} and Yang et al.'s examination of Pay-to-Win mechanics \cite{zhao2024mmo}, offers valuable insights. Yet, these studies focus exclusively on isolated scenarios and specific mechanisms within game economic systems, rather than aiming for a comprehensive game system simulation. Therefore, their systems are simplified by overlooking the interconnected consequences of player behaviors, e.g., the battle outcomes may affect purchase decision. Furthermore, these approaches are validated through qualitative observations rather than numerically validated against real-world in-game data, limiting their sufficiency for providing reliable prediction and guidance in complex game numerical design. Inspired by these works, we for the first time developed a scalable generative agent-based simulation system aiming for MMO games design. Through supervised fine-tuning (SFT) and reinforcement learning (RL) on extensive real player behavioral data, we adapt LLMs from general priors to MMO game-specific domains, enabling realistic and interpretable player decision-making. The fidelity of simulated player behaviors is validated across multi-dimension statistical data. The key contributions of this study are summarized as follows:

\begin{itemize}[leftmargin=*]
    \item MMO game simulation environment: We build a scalable system that faithfully reconstructs \textit{extraction shooter} MMO environments and incorporates a systematic intervention module. 
    
    \item High-fidelity player agents via RL: We demonstrate a framework for adapting LLMs on real behavioral data via SFT and RL to build high-fidelity player agents. These agents achieve strong alignment with ground truth across metrics including battle outcomes, item purchases, and reactions to interventions.
    
    \item Systematic validation: We conduct systematic validation using hundreds of agents over simulated weeks, enabling macroscopic monitoring of statistical distributions and microscopic analysis of individual agent behavior at every timestep, as shown in Fig.1. 
    
    \item Interpretability: The system critically provides interpretable insights for game design optimization through detailed agent behavior and reasoning analysis, moving beyond the "black-box" nature of conventional statistical simulations.
\end{itemize}

%% file: Sections/System.tex
\section{System}
\begin{figure}[tbp]
  \centering
  \includegraphics[width=0.48\textwidth]{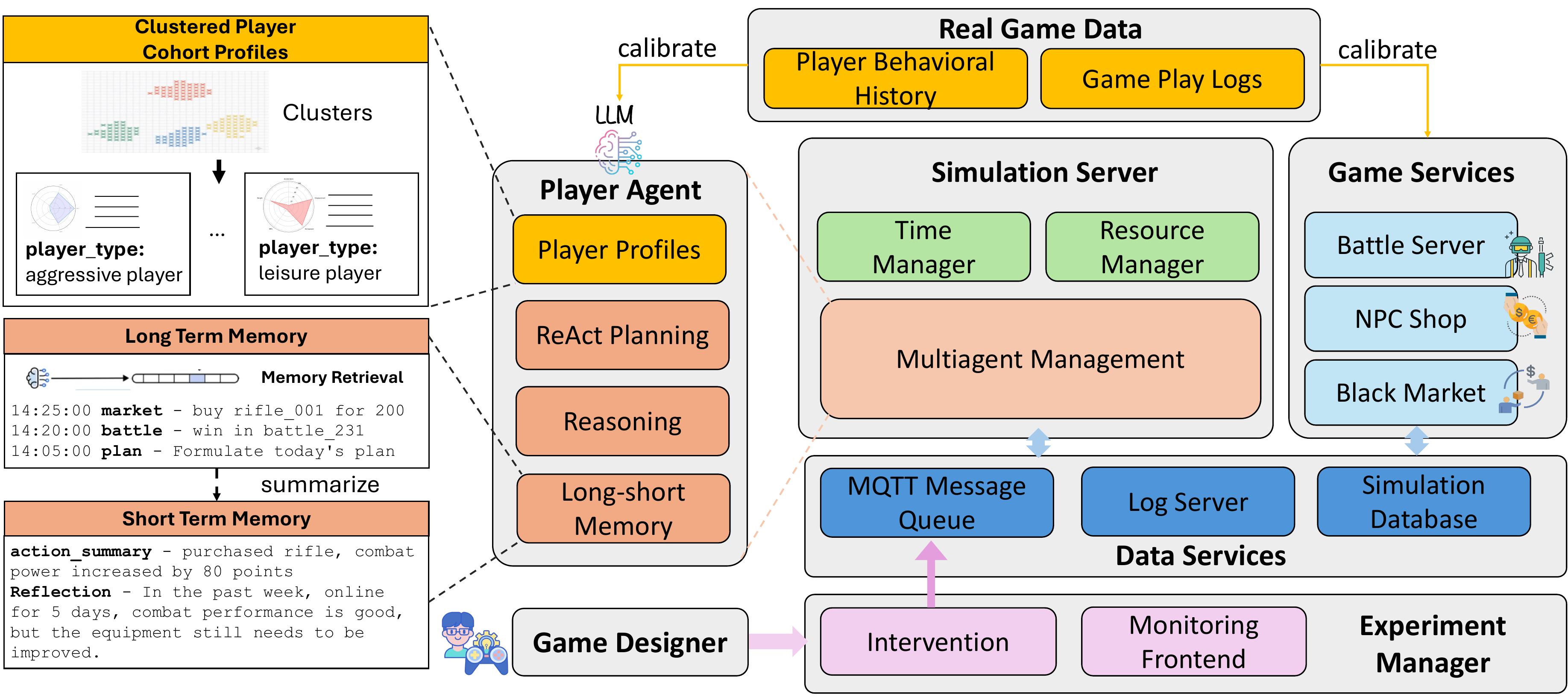}
  \caption{An overview of our agent-based game simulation system, which comprises five major components: the \textbf{Simulation Server}, \textbf{Game Services}, \textbf{Data Services}, \textbf{Experiment Manager}, and \textbf{Real Game Data}, jointly support the end-to-end workflow of large-scale, data-driven game simulations.}
  \vspace{-5ex}
  \label{fig:system}
\end{figure}

\textbf{Real Game Data.}
Real game data are stored and managed by an in-house data warehouse service. The repository contains millions of gameplay records from real players, including login and logout events, battles, purchases, and social interactions. This dataset provides rich behavioral signals that we use to construct comprehensive player profiles summarizing each player’s traits and playstyle. The historical logs are also used to calibrate both the game service models and the LLM-based player planners, ensuring that the simulated environment remains aligned with real-world dynamics.

\textbf{Experiment Manager.} 
The Experiment Manager serves as the primary interface for game designers. It provides a control panel for simulation configuration, intervention, and real-time monitoring through a graphical user interface (GUI).  
As illustrated in Fig.~\ref{fig:gui}(b), the GUI consists of four panels. The bottom panel displays a configurable simulation timeline, enabling designers to inspect the system state at any point in time. The middle-left panel lists the available agent states; selecting a state reveals all agents currently in that state for fine-grained inspection. Clicking on an individual player opens detailed attribute and history views on both sides of the interface. The middle-right panel can be used to monitor global statistics such as wealth distribution, rank distribution, resources consumption, and activeness across the entire cohort.

\textbf{Simulation Server.} 
The Simulation Server orchestrates the entire simulation lifecycle and consists of three key modules: the \textbf{Time Manager}, \textbf{Resource Manager}, and \textbf{Multi-Agent Manager}. The Time Manager maps real-world time to simulation time, allowing flexible acceleration or synchronization of virtual events. Each simulation day is discretized into $N$ time steps, during which agents transition between states and execute asynchronous tasks (e.g., battles) based on their planning decisions and the underlying state graph.  
We implement the system using Python’s asynchronous framework, enabling coroutine-based task execution for high scalability and flexible agent lifecycle control. The Multi-Agent Manager runs each agent’s routine independently so that planning and action execution are non-blocking.  
As the number of agents scales up, excessive I/O to web servers and databases can strain system resources. To mitigate this, the Resource Manager maintains a controlled pool of outbound communications, ensuring efficient resource reuse and improving system robustness and availability.

\textbf{Game Services.}
We implement three core game services that players primarily interact with: the \textbf{Battle Server}, \textbf{NPC Shop}, and \textbf{Black Market}.  
The Battle Server functions as a multiplayer arena where players engage in combat and earn in-game currency. It serves as the primary source of currency generation. The NPC Shop and Black Market, by contrast, act as currency sinks. In the NPC Shop, players purchase items directly from non-player characters (NPCs), and the spent currency is immediately recycled back into the system. 
The Black Market enables direct player-to-player trading, which use transaction taxes remove currency from circulation. Although it enhances resource exchange efficiency, it also introduces potential macroeconomic instability. Misconfigured tax rates, for example, can lead to runaway inflation or deflation, emphasizing the importance of accurate system calibration.

\textbf{Data Services.}
This layer facilitates communication, data logging, and storage during simulation.  
We employ the lightweight MQTT~\cite{Light2017Mosquitto} message queue protocol as the communication backbone connecting agents, the Experiment Manager, and the Simulation Server. MQTT supports three message types: \textit{point-to-point}, \textit{group}, and \textit{broadcast}.  
Agents use point-to-point messages for private communication (e.g., direct chat), while group channels simulate team or party chat interactions. Broadcast channels are primarily used for system-wide notifications—such as the start of in-game events, seasonal festivals, or parameter updates—that can immediately influence agents’ decision-making processes.  
A dedicated \textbf{Log Server} collects runtime logs from both the Simulation Server and Game Services. The \textbf{Simulation Database} stores all generated data, including behavioral histories, battle outcomes, and login flows. 


%% file: Sections/Method.tex
\vspace{-2ex}
\section{Simulators}
\textbf{Overview.} A faithful simulation hinges on the realism of its data-driven models. We construct two dedicated simulation modules that collectively emulate the key aspects of a game ecosystem: user behavior and battle experience. Our central focus is the Player Agent, which models how a human player plans and executes actions according to their personality, habits, and gameplay trajectory. Complementing this, the Battle Server models environmental feedback by predicting multiplayer battle outcomes using real-world data. Note that, in our scenario, black market and NPC shop are designed as deterministic systems, as commodity transactions follow predefined mechanics without stochastic elements.

\begin{figure}[tb]
  \raggedleft
  \includegraphics[width=0.48\textwidth]{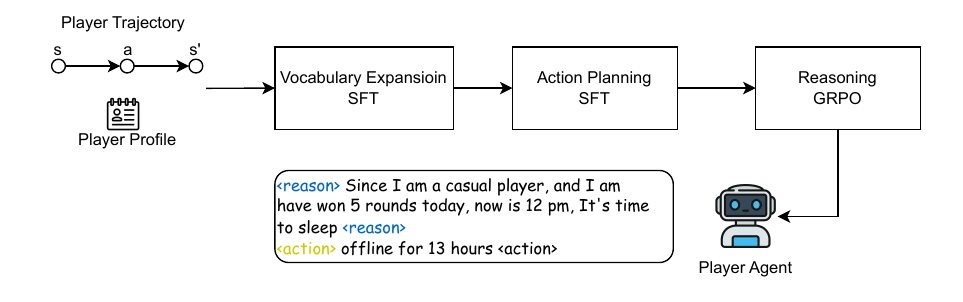}
  \vspace{-4ex}
  \caption{Player Agent Fine-tuning Pipeline. The three steps fine-tuning enables the player agent to fully adapt to the game domain and make reasonable decision like real players.}
  \vspace{-7ex}
  \label{fig:player_agent}
\end{figure}
\textbf{Player Agent.}
We adopt a three-stage fine-tuning process to endow the player agent with realistic reasoning and decision-making capabilities.
\textbf{Stage 1: Vocabulary Expansion.}
Open-source LLMs often fail to recognize domain-specific entities (e.g., “AWM”), leading to semantically inconsistent outputs. To address this, we extend the tokenizer and train the new token embeddings via SFT. We automatically generate game-specific question–answer pairs using DeepSeek-V3 and optimize a LoRA adapter with cross-entropy loss. 
\textbf{Stage 2: Action Planning SFT. }
In this stage, the model learns to predict the player’s next action—\textit{offline}, \textit{battle}, \textit{buying}, or \textit{selling}—given historical actions, environmental feedback, and player profiles. Real human decisions in games are shaped by personality traits, gameplay outcomes, and motivation (e.g., continuing after a win or logging off after a loss). We replicate such behavior patterns to enable the agent to make realistic, human-like decisions. 
\textbf{Stage 3: RL Enhancement.}
After SFT adaptation, the agent is further refined through GRPO to strengthen its reasoning capability and generalization. The RL objective encourages the agent to reason through its choices before committing to an action, aligning with how humans deliberate. This step improves both the fidelity and explainability of the simulated player’s decision process. An example of player's reflection is presented as given in Fig.1(b) Agent Details.

\textbf{Battle Server.}
In our simulated MMO game, players transit among four primary states: \textit{offline}, \textit{online}, \textit{market}, and \textit{battle}. When a player enters the \textit{battle} state, the Battle Server outputs both the win/loss result and the match income. These outcomes strongly influence subsequent behavior transitions. For example, competitive players are likely to purchase better gear after a defeat, while casual or novice players with lower frustration tolerance may choose to log off. To capture such dependencies, we train the Battle Server on large-scale consecutive match logs spanning multiple seasons from a real game. This enables the system to predict realistic win/loss outcomes and match earnings for future simulated seasons and across representative player groups, thereby ensuring that the emergent gameplay dynamics remain grounded in empirical patterns.

%% file: Sections/Experiment.tex
\section{Experiment}


\subsection{Player Agent}
\begin{figure*}[tbp]
  \centering
  \includegraphics[width=\textwidth]{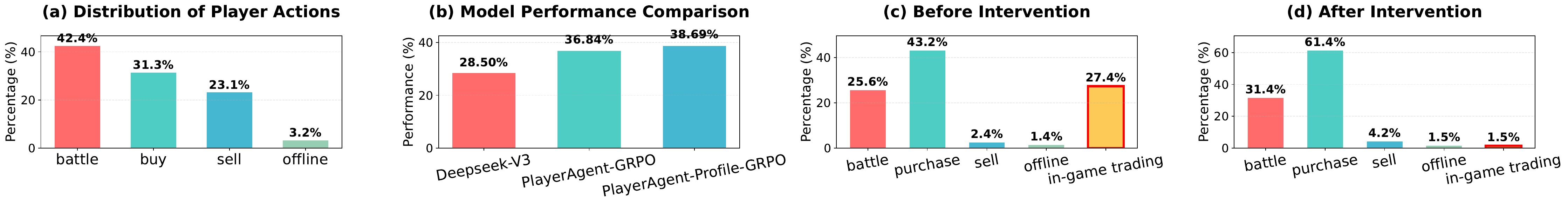}
  \vspace{-2em}
  \caption{Data distribution and accuracy.}
  \label{fig:distribution_accuracy}
  \vspace{-1em}
\end{figure*}

To verify the planning capability of our player agents, we evaluate their \textit{next-step prediction accuracy} against real player trajectories. We collected 10{,}000 gameplay trajectories from players with diverse profiles within a single day.  
At each time step $t$, a player $u$ selects an action $a_{u,t} \in \{\textit{offline}, \textit{battle}, \textit{buy}, \textit{sell}\}$, forming a four-class classification problem. The class distribution is naturally imbalanced (e.g., \textit{offline} occurrences are fewer than \textit{battle}), as shown in Fig.~\ref{fig:distribution_accuracy}(a).  

We define the stepwise prediction accuracy as:
$
    \mathrm{Accuracy} = 
    \frac{\sum_{u}\sum_{t} \mathbb{I}\!\left[f(u, t \mid x) = a_{u,t}\right]}
         {\sum_{u}\sum_{t} 1},
$
where $f(\cdot)$ denotes the player agent model, $x$ represents the contextual state information (e.g., user profile, interaction history), and $\mathbb{I}[\cdot]$ is the indicator function. For model training, we employ \textbf{Qwen2.5-1.5B} as the base language model, fine-tuned with \textbf{LoRA} adapters using rank = 16 and $\alpha = 0.2$.  

The experimental results are shown in Figure~\ref{fig:distribution_accuracy}(b). Compared to the untrained state-of-the-art baseline \textbf{DeepSeek-V3}, the model fine-tuned on user trajectories achieves a \textbf{+8.34\%} accuracy improvement. Incorporating user profile information further increases accuracy by an additional \textbf{+1.85\%}, achieving a total improvement of \textbf{+10.19\%}.  
These results demonstrate that domain-specific fine-tuning enhances the agent’s ability to reason about and predict player actions, validating the effectiveness of our adaptation pipeline.

\subsection{Battle Server}

We collect match data of 2025 \textit{S1} season, where players play within 35 and 40 matches (similar activeness). Players were clustered based on over a dozen key in-game features, including number of games, reputation, segment, playtime, game mode preference, and average killing count, etc.. This process identified and delineated five representative player profiles for the target MMO game: I.\textit{Stable Development Players}; II. \textit{Novice Players}; III. \textit{Wealth-Accumulating Elite Players}; IV. \textit{Casual Players}; V. \textit{High-skill Players}. We trained classification and regression prediction models to predict the win/loss outcome and the in-match income, respectively. The Battle Server is then validated on data in subsequent 2025 \textit{S2} season, strictly ensuring no data leakage. We aggregate 2025 \textit{S2} player data of the same five profile as the ground truth(GT), and the pred-vs-GT results of the \textit{N-th} match on 5 respective player clusters are given in Fig. 5 (b) and (c), respectively. The results indicate that the Battle Server's predictions for the win/loss accuracy and per-match income for all the five typical player groups in the new season are very accurate especially for \textit{Wealth-Accumulating Elite Players} and the \textit{Stable Development Players}. Whereas the performance of the Novice and Casual Player groups exhibits greater fluctuation, leading to comparatively larger prediction errors.

\begin{figure}[tbp]
  \centering
  \includegraphics[width=0.49\textwidth]{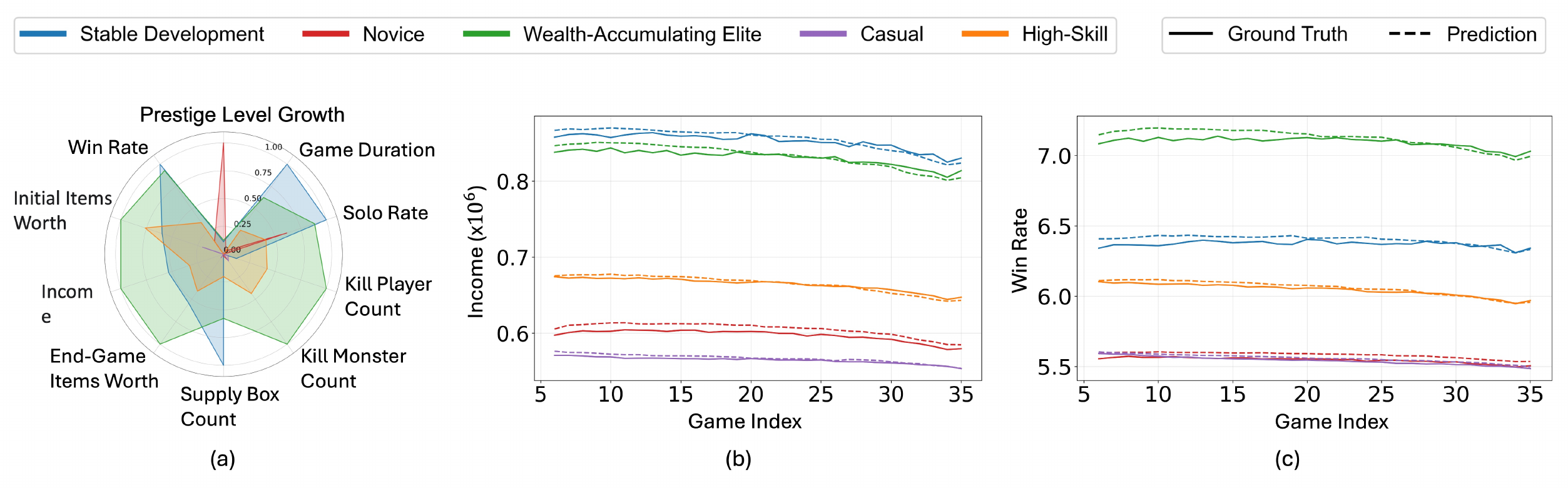}
  \vspace{-2em}
  \caption{(a) five main clusters of real players' profile; (b) and (c) show that the average income (RMSE= 0.0077) and winning rate (RMSE = 0.0016) of \textit{N-th} match on 5 respective player clusters respectively align well with real game data (with a fixed shift due to data privacy)}
  \vspace{-1em}
  \label{fig:profile_winrate_income}
\end{figure}

\subsection{Case Study of Intervention}

We conducted a case study to evaluate the efficacy of our simulation system in capturing the causal effects of external interventions.  
In the original game setting, players were not allowed to perform direct item exchanges. Consequently, to meet their demand for trading, players often resorted to informal in-game item transfers. Such spontaneous trading activities relied entirely on mutual trust between players, frequently leading to cheating or fraud.  
To mitigate this issue, the game developer later introduced an official item-exchange platform (the \textit{Black Market}) that allowed secure and regulated trading among players. This intervention was intended to both satisfy players’ trading needs and reduce fraudulent behaviors.  
In our simulation, we reproduced this real-world intervention by introducing the Black Market through the intervention model described in Fig.~\ref{fig:system}.  
Fig.~\ref{fig:distribution_accuracy}(c) and (d) compare player behavior before and after the intervention. We observe a substantial drop in the proportion of informal in-game trading, from \textbf{27.4\%} to \textbf{1.5\%}. This shift indicates the simulated player agents recognized the presence of the new, more secure trading channel and adopted it accordingly. Very few agents continued using the in-game trading mechanism, primarily due to habitual behavior formed prior to the intervention.  

These results demonstrate that our system can faithfully reproduce real-world causal effects of gameplay interventions, highlighting its utility for evaluating design decisions before deployment.